# Classifying Vietnamese Disease Outbreak Reports with Important Sentences and Rich Features

Son Doan
National Institute of Informatics
Tokyo, Japan

doan@nii.ac.jp

Nguyen Thi Ngoc Vinh
Department of Computer Science
Posts & Telecom. Institute
of Technology, Hanoi, Vietnam
vinhntn@ptit.edu.vn

Tu Minh Phuong
Department of Computer Science
Posts & Telecom. Institute
of Technology, Hanoi, Vietnam
phuongtm@ptit.edu.vn

## ABSTRACT

Text classification is an important field of research from mid 90s up to now. It has many applications, one of them is in Web-based biosurveillance systems which identify and summarize online disease outbreak reports. In this paper we focus on classifying Vietnamese disease outbreak reports. We investigate important properties of disease outbreak reports, e.g., sentences containing names of outbreak disease, locations. Evaluation on 10-time 10-fold cross-validation using the Support Vector Machine algorithm shows that using sentences containing disease outbreak names with its preceding/following sentences in combination with location features achieve the best F-score with 86.67% - an improvement of 0.38% in comparison to using all raw text. Our results suggest that using important sentences and rich feature can improve performance of Vietnamese disease outbreak text classification.

## 1. INTRODUCTION

Text classification is an important research field from mid 90s up to now. It is defined as to assign a given document into one or more predefined categories. It has many real-life applications, one of them is in web-based biosurveillance systems which identify disease outbreak reports from newswires in order to give early warnings to public health experts or governments about incoming epidemics. It plays a key role in current online Web-based biosurveillance systems such as BioCaster (http://born.nii.ac.jp), HealthMap (http://www.healthmap.org), EpiSpider (http://www.epispider.org), or GPHIN (http://www.gphin.org).

Most systems focus only in major languages such as English, Spanish, Chinese. BioCaster is the only system that can work with Asia-Pacific languages such as Vietnamese, Thai, Korean. It is shown that the ability to work with local languages is one of the key factor in providing timely warnings because local reports usually arrive before they are translated into English or other languages. Therefore there is a need to investigate the text classification of local languages. In this paper we are concerned with classifying Vietnamese disease outbreak reports.

A typical web-based biosurveillance works by examining newswire articles, finding news about disease outbreaks, extracting information about diseases and location where they are observed, and generating warning reports in a timely manner. In order to make reliable warning reports, it is crucial for biosurveillance systems to accurately differentiate news that report actual disease outbreaks from other, irrelevant ones. This problem is often difficult due to the fact that there are many news relevant to disease or health domain but not about disease outbreaks. For example, news about vaccination campaigns, health education, public health policies may contain mentions about diseases although they do not report concrete outbreaks. Such factors make it important to study the effectiveness of text classification techniques, when used with biosurveillance systems.

In recent years, there have been many studies on text classification in general [18,20], or on semi-structured texts [15], and XML classification [21]. Other researches have investigated the contribution of linguistic information in the form of synonyms, syntax, etc. in text representation [9,17] or feature selection [18,8]. Research on text classification mainly focused on two aspects: 1) feature selection, and 2) classification algorithms. As pointed out in [5,7], the use of conventional approaches in the classification process, i.e., bag-of-word, inevitably fails to resolve many subtle ambiguities, for example semantic class ambiguities in polysemous words like "virus," "fever," "outbreak," and "control," which all exhibit a variety of senses depending on the context. These different senses appear with relatively high frequency in press news reports, especially in headlines where context space is limited and creative use of language is sometimes employed to catch attention. A further challenge is that diseases can be denoted by many variant forms. Therefore, we consider that the use of advanced natural language processing (NLP) techniques like named entity recognition (NER) and anaphora resolution are needed in order to achieve high classification accuracy.

In research context of disease outbreak text classification, there is a challenge for studies of local languages: data sets for training and evaluation of classification systems are usually not available for public. Another challenge is how to improve classification performance. In recent years, several studies have addressed this problem. For example, Doan et al. [6,7] investigated the roles of semantic roles and named entities. They showed that such features were beneficial for the system's accuracy. Conway et al. [4] studied n-grams, semantic features, and automated feature selection methods. The authors reported that removing irrelevant features could lead to substantial improvement in classification accuracy. For Vietnamese text classification, Hoang et al. [11] showed that n-gram model was beneficial to other models. In all these studies, features are extracted from all content of news. In addition to feature selection, it has been shown that different components of a news report such as title, first paragraph, last paragraph contribute differently to the classification accuracy [7].

In this paper, we make a step further and study the usefulness of text surrounding disease names in classifying the whole news as reporting a disease break or not. We conjecture that only a few words located near a disease name are informative enough to classify the whole news with respect to that disease. For example, considering only phrase "từ đầu năm đến nay đã có 2 người ở Sóc Trăng, Kiên Giang tử vong vì nhiễm cúm A/H5N1" (since the beginning of the year there have been 2 people in Soc Trang, Kien Giang died of A/H5N1) excerpted from a news report, is sufficient to conclude that the report is relevant without reading the whole report.

We experimentally verify the hypothesis that a few important words/sentences in news reports provide enough information to achieve the same or higher level of classification accuracy than the whole texts do. We also point out which sentences give the most accurate results when used as features for classifiers. These findings have a number of applications. First, they can be used to improve classification accuracy. Second, they can be used for text summarization, where the few important sentences summarize the news up to the extent that they provide the same classification power. And third, these findings make it possible to classify news reports each containing different disease names.

Our main contributions in this paper are two-fold: 1) we developed an annotated corpus for Vietnamese disease outbreak reports; 2) we showed that using only few but important sentences in combination with location information can improve performance of text classification. Those findings might have potential applications in the real-life Web-based biosurveillance systems.

## 2. METHOD

### 2.1 Main idea

The main goal of this paper is to verify the hypothesis that in classifying newswire articles into disease outbreak relevant and irrelevant ones, using a few but important sentences with rich features can lead to performance as good as using all information from the whole text. As the example in Introduction shows, a disease outbreak report often contains one or two sentences mentioning about the name of disease outbreak and the location where it happens. It comes to us that sentences containing disease names and location information are two key features in outbreak reports. In addition, we investigate the roles of surrounding sentences to the outbreak sentence as well as additional features such as titles or reports mentioning two or more outbreak diseases.

### 2.2 Corpus

The corpus contains 1,544 Vietnamese reports which 762 are positive and 782 are negative reports. Vietnamese news reports were collected from main Vietnamese online news providers such as Vnexpress (http://vnexpress.net), VietnamNet (http://www.vietnamnet.vn), Tuoi tre (http://tuoitre.vn) and Google News Vietnam (http://news.google.com.vn/). Based on the BioCaster guideline [14], we developed a simple guideline for annotation. The guideline categorizes four types of reports: *alert*, *check*, *publish* and *reject*. We grouped *alert*, *check* and *publish* as the *relevant* class and *reject* as the *non-relevant* class [14]. In general, relevant reports are reports talking about first case reported or warning about spreading of an outbreak disease. Non-relevant reports can be divided into two groups: reports not mentioning about disease outbreak, and reports mentioning disease outbreak but are not related to any epidemic or pandemic such as about vaccination campaigns, health education, and public health policies.

This is an example of *relevant* Vietnamese disease outbreak reports with tagged disease names and locations.

```
<ner type=disease>Cúm A/H1N1</ner> lan rộng

Cục Y tế dự phòng (Bộ Y tế) cho biết, <ner
type=disease>cúm A/H1N1</ner> đã lan rộng ra 35
tỉnh, thành trong cả nước, trong đó có hơn 200 ca
mắc dịch, 7 trường hợp đã tử vong.

Mới nhất, ngày 1/4, Sở Y tế tỉnh Bến Tre, cho biết
trên địa bàn tỉnh vừa xuất hiện một ổ dịch mới
<ner type=disease>cúm A/H1N1</ner> tại xã Tân
Phong, huyện <ner type = location> Thạnh Phú
</ner>.
```

The corresponding English translation is given below.

```
<ner type=disease> A/H1N1 flu </ner> are spreading

The Preventive Health Agency (Ministry of Health)
reported that <ner type=disease> A/H1N1 flu </ner>
has spread into 35 provinces and cities in the
country, including more than 200 infected cases, 7
cases were died.

At the earliest, on April 1st, The Health
Department of Ben Tre province, said that on the
province there has been a new outbreak of <ner
type=disease>A/H1N1 flu </ner> at Tan Phong
village, <ner type=location> Thanh Phu </ner>
district.
```

Two graduate students annotated the corpus, and one of the authors checked again to confirm the consistency of annotation. There are two challenges in building the good corpus in order to build a good classifier: 1) balance data between relevant and non-relevant reports, and 2) the corpus should cover as much as possible the number of difficult reports which are "borderline" between relevant and non-relevant class. To overcome these, we searched within news providers using several keywords related to outbreak diseases such as "bệnh cúm" (flu), "A/H1N1", "dịch bệnh" (outbreak) to get more disease outbreak reports.

Statistics of corpus and information about its features are shown in Table 1.

**Table 1. Statistics about Vietnamese disease outbreak corpus.**

| Names | Relevant | Non-relevant |
|---|---|---|
| # Reports | 762 | 782 |
| # Disease outbreak sentence | 2,652 | 1,254 |
| # Disease outbreak sentence + the preceding sentence | 1,086 | 527 |
| # Disease outbreak sentence + the following sentence | 1,747 | 818 |
| # Disease outbreak sentence + the preceding and following sentences | 771 | 500 |
| # Reports which have location features | 486 | 153 |

### 2.3 Features and models for text classification

We investigate 14 models with corresponding features as follows (note that in all models, we use unigrams with the traditional bag-of-word representation).

1. Baseline: Raw text only, i.e., all text in news are used as features.

Next, we consider the effects of sentence containing disease outbreak and its preceding/following sentences. Only such sentences are used as features in representation for reports, including:

2. Sentences containing disease names. Only sentences containing disease names are used as features.

3. Sentences containing disease name + the preceding sentence: Only sentences containing disease names and their preceding sentence are used as features. If there is no preceding sentence it will leave as a blank.

4. Sentences containing disease name + the following sentence: Sentences containing disease names and its following sentence are used as features. If there is no following sentence it will leave as a blank.

5. Sentences containing disease name + preceding and following sentences: Sentences containing disease names and its preceding and following sentences are used as features. If there is no preceding or following sentence it will leave as a blank.

We then consider about the effects of location feature, i.e., if sentences in consideration contain location name, it will be assigned as 1, otherwise 0. The details are described as follows.

6. Sentences containing disease name + the preceding sentence + location feature: Sentences containing disease names, its preceding sentence and location are used as features. If such sentences containing a location name then location features are assigned into 1, otherwise 0.

7. Sentences containing disease name + the following sentence + location feature: Features are sentences containing disease names, its following sentence and location features.

8. Sentences containing disease name + preceding and following sentences + location feature: Features are sentences containing disease names, its preceding/following sentences and location features.

We also consider the importance of title in representation as follows.

9. Sentences containing disease name + the preceding sentence + location feature + title: Features are sentences containing disease name, its preceding sentence, location feature and the title.

10. Sentences containing disease name + the following sentence + location feature + title: Features are sentences containing disease name, its following sentence, location feature and the title.

11. Sentences containing disease name + preceding and following sentences + location feature + title: Features are sentences containing disease name, its following/ preceding sentences, location feature and the title.

Finally we consider the importance of reports talking about multiple disease names, i.e., if reports talking about more than two disease names it will be assigned as 1, otherwise 0. Also, in such cases sentences surrounding all disease names are used to extract features. The corresponding models are as follows.

12. Sentences containing disease name + the preceding sentence + location feature + title + multiple disease feature.

13. Sentences containing disease name + the following sentence + location feature + title + multiple disease feature.

14. Sentences containing disease name + preceding and following sentences + location feature + title + multiple disease feature.

## 2.4 Classification Algorithms.

We used the SVM algorithm for classification since it has been shown as the state-of-the-art method in text classification [19,12,5,4]. For text representation, we used binary features, i.e., a feature can take on a value of 1 if it presents and 0 otherwise. No stemming or stop words removal step was applied.. We chose the SVMlight package [16] and the linear kernel with *c-value* as 1.0 as parameters in the SVM algorithm which are the same as used by Doan et al. [5] and Conway et al. [4].

## 2.5 Evaluations

We used traditional Precision/Recall measures and F-score to evaluate the effectiveness of each model. Precision and Recall are defined as follows:

Precision = TP/(TP+FP),

Recall = TP/(TP + FN),

where:

TP is the number of relevant documents correctly classified as relevant.

FP is the number of irrelevant documents incorrectly classified as relevant.

FN is the number of relevant documents incorrectly classified as irrelevant.

F-score is calculated as:

F-score = 2*Precision*Recall/(Precision + Recall)

In this paper, we used 10-times 10-fold cross-validation which was described in Bouckaert and Frank [1] to compare different models under test. This procedure runs 10 times, each split training/test data with a different random seed. It will result 100 training/test split for each model. In the next section we report results averaged over 100 runs.

## 3. RESULTS AND DISCUSSION

### 3.1 Results

Main results for 10-times 10-fold cross-validation are shown in Table 2. The baseline method that used full text achieved an F-score of 86.29%, while the model using only a single sentence that contains disease name achieved an F-score of 85.53%. These results show that a single sentence containing disease name can provide classification accuracy very close to that of full text. Adding previous/following sentences as features increased F-measure to 85.53%. by using sentence containing disease names and increased to 85.83% by adding previous/following sentences. This is very close to using all raw text, though the performance is still slightly behind that of the baseline. When we added location features the performance substantially improved, increasing from 85.12% to 85.96% F-score for sentences containing disease with the previous sentence (model #3), 85.48% to 86.27% F-score for sentences containing disease with the following sentence (model

#4) and from 85.83% to the best F-score of 86.67% for sentences containing disease with the previous/following sentences (model #5). These results also show that surrounding sentences of those containing disease outbreaks also contribute to the performance of text classification. In addition, it shows that the sentence immediately following the sentence with disease contributes more to classification accuracy than the one that comes before. Indeed, models with following sentences all achieved higher F-scores than models with preceding ones: 85.48% F-score of model #4 vs. 85.12% F-score of model #3, and 86.27% F-score in model #7 vs. 85.96% F-score in model #6

Models #9,10,11 in Table 2 show effects of title and models #12,13,14 in Table 2 show effects of multiple disease features on classification accuracy with the presence of sentences containing outbreak diseases. Interestingly, both title and multiple disease features show negative effects, i.e., they tend to reduce performance of classification. For example, it reduces from 86.67% F-score to 86.55% F-score with title and 86.50% F-score with multiple disease features. It shows that the presence of sentences containing disease outbreak names and location features might be efficient enough for achieving the best F-score.

We also observed that the size of vocabulary used in proposed models is much smaller than that of the baseline. Specifically, the size of vocabulary used in the baseline model is 10,101 while all remaining models use vocabularies with less than 5,000 words. This gives the proposed models an advantage over using full text in terms of computational cost. This advantage is very important for systems working in the real scale of online newspapers medium.

**Table 2. Results on 10-times 10-fold cross-validation on the Vietnamese disease outbreak corpus.**

| # | Models | #Vocabulary | F-score (%) |
|---|---|---|---|
| 1 | Baseline (raw text) | 10,101 | 86.29 |
| 2 | Sentence containing disease name | 3,997 | 85.53 |
| 3 | (2) + the preceding sentence | 4,211 | 85.12 |
| 4 | (2) + the following sentence | 4,316 | 85.48 |
| 5 | (2) + preceding and following sentences | 4,463 | 85.83 |
| 6 | (3) + location feature | 4,211 | 85.96 |
| 7 | 4) + location feature | 4,316 | 86.27 |
| **8** | **(5) + location feature** | **4,463** | **86.67** |
| 9 | (6) + title | 4,703 | 85.80 |
| 10 | (7) + title | 4,810 | 86.22 |
| **11** | **(8) + title** | **4,936** | **86.55** |
| 12 | (9) + multiple disease feature | 4,703 | 85.75 |
| 13 | (10) + multiple disease feature | 4,810 | 86.22 |
| **14** | **(11) + multiple disease feature** | **4,936** | **86.50** |

## 3.2 Discussion

There are several interesting findings from the results in Table 2. First, the best performance we achieved by using the outbreak sentence and a +/-1 sentence window and location information. It indicates that the most important information for classifying a news report tend to concentrate in a few sentences near disease names, and they are a key factor to identify relevant reports. Second, between two sentences that precedes and follows the sentence with a disease name, the following is more important than the preceding one. It indicates that the text following the outbreak sentence seems to be more relevant to disease outbreak than the preceding one. A possible explanation for this phenomenon is that the text immediately following the sentence with disease name is often used to detail about what has been said about the disease, and thus provide necessary information for classifiers. After a closer look at the data, we found that following sentences often contain details about the potential of epidemic happening after that or reported about possibilities of spreading of disease outbreak. Third, given information of outbreak sentences and location information, title has a negative effect. This is a surprising finding if compared with previous results: Doan et al. [7] reported that title is the most important section within the whole report, our results show that the outbreak sentence is more important. Fourth, multiple disease feature tend to reduce performance, it can be interpreted that reports talking about more than two disease outbreaks often are general reports than alerting reports. In some cases, a news containing multiple disease may report about an outbreak of one disease but not of the other ones.

*Comparison among sentences and rich features*

From the obtained results, we can rank the importance of sentence as follows: outbreak sentence >= the following sentence >= the preceding sentence and rich features are as follows: location feature >= title >= multiple disease features. These rankings might be useful indicators for identifying Vietnamese disease outbreak reports.

*The potential application in Vietnamese disease outbreak report classification system.*

It can be questioned that how our results can be applied into real-life applications. The finding in the results show a practically potential application in a biosurveillance system: we can use several sentences within news reports with rich features for text classification instead of using the whole text. More specifically, if we can identify sentences containing disease and location names, we can build a classification system with high performance. We note that in real Web-based biosurveillance systems, reports have to be pre-processed before processing text. This is because Internet reports are often in HTML, XML formats or any un-structured format with many meta or embedded data such as figures, tables, captions or advertisements, thus such noise data need to be cleaned. In other scenario, we observed that performance to identify disease outbreak and location names in Vietnamese reports often very high with about 90% F-score by using SVM in combination with rule-based methods. These shows that it is possible to build a simple but efficient system for Vietnamese disease outbreak report classification system.

## 4. CONCLUSION

In this paper we investigate the roles of few but important sentence and rich features in Vietnamese disease outbreak text classification. We have built an annotated corpus for Vietnamese disease outbreak reports. The results have shown that using only outbreak sentence with its previous and following sentences in combination with location information can achieve better performance than using raw text. In the future we plan to extend our methods into English disease outbreak corpus.

## 5. ACKNOWLEDGEMENT

Funding for this study was provided by Ministry of Science and Technology of Vietnam, Program KC.01/11-15.